\pdfoutput=1
\documentclass[11pt]{article}

\usepackage[]{acl}

\usepackage{times}
\usepackage{latexsym}

\usepackage[T1]{fontenc}
\usepackage[utf8]{inputenc}

\usepackage{microtype}

\usepackage{graphicx,xcolor}
\usepackage{pxrubrica}   
\usepackage{url}
\usepackage{booktabs}
\usepackage{multirow}
\usepackage{xspace}
\usepackage{amsmath}
\usepackage{subfigure}
\usepackage{comment}
\usepackage{makecell}
\usepackage{color}
\usepackage{pifont}
\usepackage{tabularx}
\usepackage{colortbl}
\newcommand{\yes}{\ding{55}}%
\newcommand{\no}{\ding{51}}%

\definecolor{red}{rgb}{0.8,0,0}
\definecolor{green}{rgb}{0,0.8,0}
\definecolor{blue}{rgb}{0,0,0.8}
\definecolor{yellow}{rgb}{0.90,0.91,0.75}
\definecolor{purple}{rgb}{0.85,0.80,0.91}

\newcommand{\rcOurs}[0]{\rowcolor[rgb]{0.85,0.98,0.87}}

\newcommand{\breakInTable}[1]{\begin{tabular}{c} #1 \end{tabular}}
\newcommand{\methodName}[0]{VisCE$^2$\xspace}
\renewcommand{\paragraph}[1]{\noindent\textbf{#1}.\hspace{2pt}}

\title{Vision Language Model-based Caption Evaluation Method\\  Leveraging Visual Context Extraction}

\author{
    Koki Maeda$^{\clubsuit, \spadesuit}$   
    Shuhei Kurita$^{\spadesuit}$ 
    Taiki Miyanishi$^{\diamondsuit, \spadesuit}$ 
    Naoaki Okazaki$^{\clubsuit}$\\
      $^{\clubsuit}$Tokyo Institute of Technology 
      $^{\spadesuit}$RIKEN AIP 
      $^{\diamondsuit}$ATR \\
      \texttt{koki.maeda@nlp.c.titech.ac.jp},
      \texttt{shuhei.kurita@riken.jp},\\
      \texttt{miyanishi@atr.jp},
      \texttt{okazaki@c.titech.ac.jp} \\}

\begin{document}
\maketitle

\begin{abstract}

Given the accelerating progress of vision and language modeling, accurate evaluation of machine-generated image captions remains critical. In order to evaluate captions more closely to human preferences, metrics need to discriminate between captions of varying quality and content. However, conventional metrics fail short of comparing beyond superficial matches of words or embedding similarities; thus, they still need improvement. This paper presents \methodName, a vision language model-based caption evaluation method. Our method focuses on visual context, which refers to the detailed content of images, including objects, attributes, and relationships. By extracting and organizing them into a structured format, we replace the human-written references with visual contexts and help VLMs better understand the image, enhancing evaluation performance. Through meta-evaluation on multiple datasets, we validated that \methodName outperforms the conventional pre-trained metrics in capturing caption quality and demonstrates superior consistency with human judgment.

\end{abstract}

\section{Introduction}
The evaluation of the machine-generated image caption is a core research topic to illustrate models' ability to describe their visual observation in textual forms and shape the branch of vision and language modeling studies into meaningful and grounded directions. In the early stage of neural image captioning research, such as neural image caption generator~\cite{ShowAndTell}, attention-based~\cite{ShowAttendTell}, and sentinel and spatial attention~\cite{Lu-etal-2017-knowing}, models had enabled more and more detailed and accurate descriptions. Hence, they achieved better and better performance on reference-based automatic evaluation metrics such as BLEU~\cite{BLEU} and CIDEr~\cite{CIDEr}. SPICE~\cite{SPICE} was also proposed to access better correspondences of reference and generated captions in grammatical aspects. More recently, BERTScore~\cite{BERTScore} and CLIPScore~\cite{CLIPScore} have been introduced and are utilized to measure the similarity of embeddings of the generated text and references.

In very recent advances in vision and language models (VLMs), however, the model ability often seems to exceed the evaluation metrics themselves or even the coverage of the set of annotated references. Both InstructBLIP~\cite{InstructBLIP} and LLaVA~\cite{LLaVA} follow textual instructions and generate tailored descriptions that are not even similar to references but are of high quality. 
We, hereby, go back to the basics of image captioning: first capture the contents of the images, then rearrange them to describe and compose a phrase.  As an auto-evaluation metric, this corresponds to accessing different abilities: coverage of image contents, accuracy for describing them, and writing composition. In this line of evaluation metrics, InfoMetIC~\cite{InfoMetIC} relies on the matching of image regions and words in the captions. While it is quite sensitive to the alignments of objects that appear in images and captions, it becomes less sensitive to supportive facts such as attributes or interactions of objects. Unfortunately, it is also a learning-based metrics and relies on fine-tuned with multiple captioning datasets.

This paper concentrates on reference-free image caption evaluation and proposes a new \underline{\textbf{Vis}}ion Language Model-based \underline{\textbf{C}}aption \underline{\textbf{E}}valuation Method leveraging \underline{\textbf{Vis}}ual \underline{\textbf{C}}ontext \underline{\textbf{E}}xtraction (\textbf{\methodName}). In this context, ``visual context'' refers to information about objects, attributes, and their relationships, including those in the background and inconspicuous objects. This visual context is given to the model in the evaluation process with an image and a candidate caption. Explicitly providing the visual context in a structured format, rather than reference images or hand-written captions, helps the VLM better understand the images' content and expose how accurately the candidate caption describes the parts of the image or which parts are missing. 

We investigated the quality of the proposed evaluation method on several image caption datasets, THumB~\cite{THumB}, Flickr8k-Expert~\cite{Flickr8k}, Composite~\cite{Composite}, and Pascal-50S~\cite{CIDEr}. Our method, \methodName, outperformed conventional metrics and was highly correlated with human judgments. Furthermore, meta-evaluation experiments uncovered that the evaluation scores of \methodName strongly correlate with the accuracy of candidate captions. In ablation studies, we verified that the proposed method is effective with the state-of-the-art VLM. Moreover, we revealed that Applying \methodName to LLaVA-v1.5~\cite{LLaVA1.5} improves evaluation performance comparable to that of GPT-4V~\cite{GPT-4} without visual context.

\section{Related Work}

\begin{figure*}[t!]
    \centering
    \includegraphics[width=\linewidth]{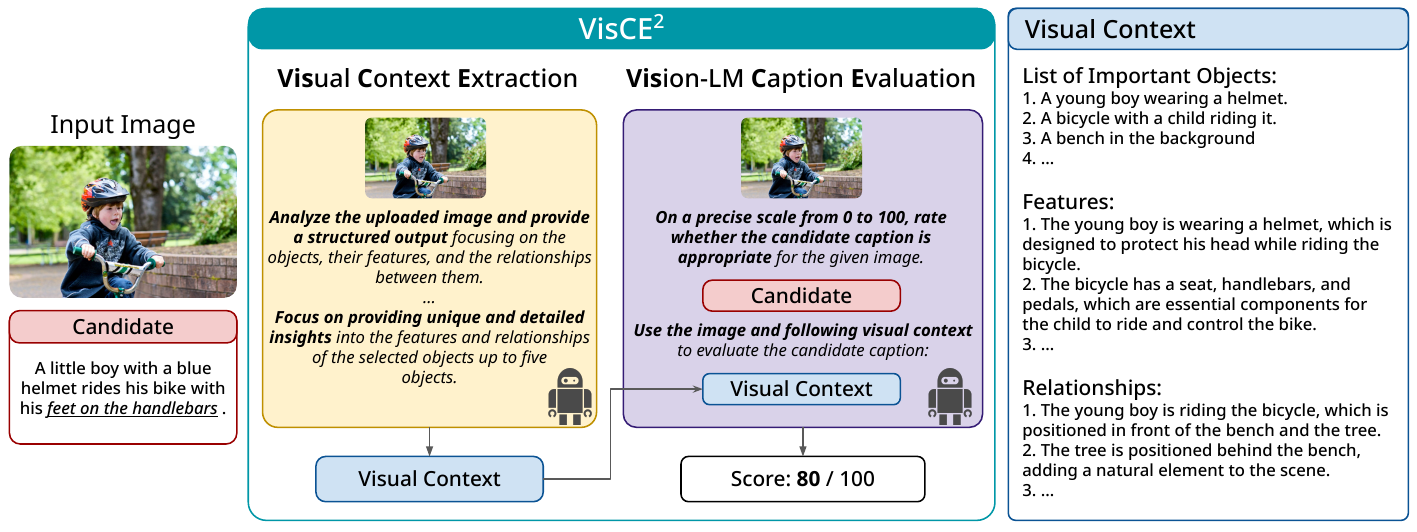}
    \caption{Overview of automatic caption quality evaluation by \methodName and example of the input/output. First, VLM extracts the visual context from the image, organized in a bullet list format, presenting objects, object features, and relationships between objects. Then, VLM evaluates the caption using the obtained visual context along with the image and candidate caption.}
    \label{fig:fig1-method-overview}
\end{figure*}

\subsection{Evaluation Method for Image Captioning}

\paragraph{Text-only (reference-based) methods}
Image captioning has been evaluated using a combination of several metrics. While some have been adapted from metrics used in other NLP tasks such as machine translation and summarization  (BLEU~\cite{BLEU}, METEOR~\cite{METEOR}, ROUGE~\cite{ROUGE}) and others proposed for image captioning (CIDEr~\cite{CIDEr}, SPICE~\cite{SPICE}), all of them are mainly based on \emph{n}-gram matches with the reference caption.
Following these classical approaches, an evaluation metric was proposed that exploits the versatility of pre-trained models: BERTScore~\cite{BERTScore} measures the similarity of embeddings output by BERT~\cite{BERT} for each reference and candidate caption. In BERTScore++~\cite{BERTScorePP}, they fine-tuned BERT to the task using the image caption dataset.
More recently, CLAIR~\cite{CLAIR} utilized large language models for evaluation.
This ensembles ChatGPT~\cite{ChatGPT}, Claude~\cite{Claude}, and PaLM~\cite{PaLM} and achieves high evaluation performance by providing only reference sentences and instructions for the task.

\paragraph{Crossmodal methods}
Covering the rich visual information in an image with only predefined reference captions is difficult. To alleviate this loss of information, evaluation methods based on vision and language models (VLMs), which leverage features of the images, have been proposed. TIGEr~\cite{TIGEr} uses a pre-trained SCAN model~\cite{SCAN}, fine-tuned on the COCO dataset~\cite{MSCOCO}, and calculates how much the candidate caption is grounded in the image. ViLBERTScore~\cite{ViLBERTScore} extracts features of the images and calculates the score as with BERTScore with a pre-trained ViLBERT model~\cite{ViLBERT}. FAIEr~\cite{FAIEr} connects images and texts via scene graphs and computes scores according to their overlap.

\paragraph{Image-only (reference-free) methods}
To simultaneously improve evaluation performance and reduce the cost of annotation for references, several methods have been proposed for image caption evaluation that apply quality estimators or pre-training models trained on large-scale web-crawled datasets.
CLIP-S~\cite{CLIPScore} attaches the score by simply calculating the modified cosine similarity between an embedding of an image and that of a candidate caption with the CLIP~\cite{CLIP}. PAC-S~\cite{PAC-S} further refined the pre-training method of CLIP with data augmentation using an image captioner~\cite{BLIP} and image generator~\cite{StableDiffusion}, resulting in improved evaluation performance. UMIC~\cite{UMIC} fine-tunes UNITER~\cite{UNITER} via contrastive loss for gold and automatically perturbated caption pair. InfoMetIC~\cite{InfoMetIC} fuses image and language modalities by stacking CLIP and Transformer~\cite{Transformer}, fine-tuned with large image-caption datasets~\cite{Flickr30k,Composite}.

Unlike existing reference-free methods, our method utilizes visual context to detail the structure of the image content for caption evaluation.
In our evaluation protocol, the VLM extracts the visual context to encapsulate the comprehensive image content and feeds it to the VLM itself. By doing so, it is possible to reference information from both the vision and language sides and hence is expected to improve the accuracy of quality estimation.

\subsection{Recent Vision-Language Models}

The recent vision and language model proposal has progressed the fusion understanding of image and language modalities. One of the most important milestones is CLIP~\cite{CLIP}. Contrastive learning on a large-scale image-text dataset constructed by web crawling has significantly improved zero-shot performance on various vision and language tasks. Following this, the development of various models accelerated progress in the V\&L domain. BLIP~\cite{BLIP} refined pre-trained methods by enhancing the quality of image and text data by the image captioner. OFA~\cite{OFA} integrated various V\&L tasks by modifying the input-output architecture. Several other studies also presented remarkable performance, which are still being upgraded~\cite{Qwen-VL,MiniGPT-4,MiniGPT-v2,BLIP-2}. In addition, InstructBLIP~\cite{InstructBLIP} and LLaVA~\cite{LLaVA} allow us to perform various tailored tasks according to instructions. Furthermore, state-of-the-art VLMs such as GPT-4V~\cite{GPT-4} and Gemini~\cite{Gemini} are now available via APIs.

We focus on the fidelity to instructions and hence applied our \methodName to the LLaVA-v1.5~\cite{LLaVA1.5}, the improved version of open-source LLaVA, to ensure transparency and reproducibility in our experiments. We also tested the evaluation performance on commercial GPT-4V to verify whether our method is effective with other VLMs and to measure the upper bound of the VLM-based evaluation methods.

\section{Proposed Method: \methodName}
We introduce a reference-free automatic caption evaluation method \textbf{\methodName} with VLM.
Figure \ref{fig:fig1-method-overview} shows the overview of \methodName.
Our method takes an image and a caption as inputs and predicts a rating score for how accurately the caption describes the image.
This assessment is conducted via two key components: visual context extraction and VLM-based caption evaluation.

\paragraph{Visual Context Extraction}
\methodName initially extracts a detailed visual context from an input image using VLM.
We define \emph{visual context} as the content of an image classified into objects, object attributes (including color, shape, and size), and relationships between objects following the similar notion of scene graphs~\cite{SceneGraph}.
To extract relevant image details, we instruct the VLM to articulate the visual context in a bullet list format by category rather than employing the structured graphical representation due to its complexity and the difficulties associated with textual representation.
We believe that providing the successive VLM module with detailed visual context in a structured format helps bridge the gap between the reference image and the candidate text. 
This facilitates the model to evaluate both consistently and comprehensively.

\paragraph{Vision-LM Caption Evaluation}
In the next step, \methodName evaluates the candidate caption based on the extracted visual context and an input image.
As with many other LLM-based tasks, \methodName treats caption evaluation as a multimodal text completion task.
The VLM is given a prompt combining a reference image, visual context, and a candidate caption as input and then generates an output sentence that encapsulates the overall quality scores ranging from 0 to 100.
In the preliminary experiments, we found that the VLM occasionally disregards instructions to generate only evaluation scores, instead producing sentences that include these scores (e.g.,  \emph{``The score is X out of 100''}.)
We eliminated these canonical phrases in the postprocessing phase and designated the first integer value in the output sentences as the evaluation score. 
In contrast to previous embedding-based methods, such as CLIP-S~\cite{CLIPScore}, which cannot determine the quality of scores without multiple examples, our approach can provide absolute scores close to human intuition, allowing for determining the caption's quality by just looking at a single caption.

\section{Experiment}
This section demonstrates the effectiveness of our \methodName and conducts a meta-evaluation of the automatic evaluation methods.

\subsection{Experimental Settings}

\paragraph{Implemntation details}
To ensure the transparency and reproducibility of experimental results, we use
\texttt{LLaVA-v1.5-13B}\footnote{\url{https://huggingface.co/liuhaotian/llava-v1.5-13b}}, the best-performing model among the publicly available VLMs with the default hyperparameter settings and report the results of a single run.
Note that \methodName allows for the use of any VLM without restrictions. 
Hence, we perform a meta-evaluation of GPT-4V in Sec. ~\ref{subsec:gpt4v} to validate this and explore the upper bounds on the evaluation performance of current VLMs.

\paragraph{Baseline metrics}
We compared the evaluation performance against five of the most common metrics in the automatic evaluation of image captioning: BLEU~\cite{BLEU}, ROUGE~\cite{ROUGE}, METEOR~\cite{METEOR}, CIDEr~\cite{CIDEr}, and SPICE~\cite{SPICE}.
These metrics evaluate the overlap of \emph{n}-grams with reference captions.
Specifically, SPICE generates semantic scene graphs from the candidate captions using dependency parse trees, assessing matches based on objects, attributes, and their relationships.
Furthermore, we employed more recent reference-based measure BERTScore++~\cite{BERTScorePP} and modern reference-free metrics, CLIP-S~\cite{CLIPScore}, PAC-S~\cite{PAC-S}. 
Among reference-free metrics, InfoMetIC~\cite{InfoMetIC} relies on a quality estimator that is fine-tuned with the training splits of Flickr-30k~\cite{Flickr30k} and MS-COCO~\cite{MSCOCO}.
In contrast, the other metrics, including our \methodName, do not depend on such data-specific fine-tuning. 
Furthermore, while our method uses a publicly available model, CLAIR~\cite{CLAIR} exploits the zero-shot evaluation results of proprietary LLMs such as ChatGPT, Claude, and PaLM.
While we include their results for comparison in the subsequent experiments, it should be noted that their results are not directly comparable without careful consideration. 

\paragraph{Measures for meta-evaluation}
In our meta-evaluation experiments, we utilized three different indicators for each evaluation dataset, following previous studies:
Pearson correlation for measuring the linear correlation between two sets of data,
Kendall’s Tau for measuring the ordinal association between two measured quantities and 
the accuracy as of the percentage of the correct pairwise ranking between two candidate captions.

\begin{table*}[t!]
    \centering
    \small
    \caption{
        The effect of textual resources given to the model on evaluation performance on multiple image caption datasets. Correlations with human ratings are measured by Pearson's $\rho$, Kendall's $\tau$, and agreement with human preference by accuracy. \textbf{Bold} means the best scores.
    }
    \label{tab:tab1-effect-visual-context}
    \scalebox{1}{
        \tabcolsep = 0.75mm
        \renewcommand{\arraystretch}{1.2}
        \begin{tabular}{lc|cccccc|c|c|c}
          \toprule
            \multirow{2}{*}{\textbf{Method}} &
            \multirow{2}{*}{\breakInTable{\textbf{Ref-}\\\textbf{free}}} & 
            \multicolumn{3}{c}{\breakInTable{\textbf{THumB w/o human} \\ Pearson's $\rho$}} &
            \multicolumn{3}{c}{\breakInTable{\textbf{THumB w/ human} \\ Pearson's $\rho$}} &
            \multirow{2}{*}{\breakInTable{\textbf{Flickr8k-Exp.} \\ Kendall's $\tau$}} &
            \multirow{2}{*}{\breakInTable{\textbf{Composite} \\ Kendall's $\tau$}} &
            \breakInTable{\textbf{Pascal-50S} \\ Accuracy (\%)} \\
            &  &  P & R & Total & P & R & Total &  &   & Mean \\
          \midrule 
             Vanilla            & \no  
                & .44 & .08 & .38 & .41 & .08 & .34 
                & 55.9  & 52.4 
                & 80.5 \\
             Reference          & \yes 
                & .42 & \textbf{.12} & .40 & .40 & \textbf{.14} & .38 
                & 54.6  &  54.5 
                & 79.2 \\
             Description        & \no  
                & .53 & .08 & .44 & .48 & .07 & .39 
                & 57.4  &  52.5 
                & 77.5 \\
             \rcOurs\textbf{\methodName (Ours)} & \no  
                   & \breakInTable{\textbf{.54} \\ \color{green}(+.10)} 
                   & \breakInTable{.08 \\ \color{gray}($\pm$.00)} 
                   & \breakInTable{\textbf{.45} \\ \color{green}(+.07)}
                   & \breakInTable{\textbf{.49} \\ \color{green}(+.08)}
                   & \breakInTable{.07 \\ \color{gray}(-.01)}
                   & \breakInTable{\textbf{.39} \\ \color{green}(+.05)} 
                   & \breakInTable{\textbf{59.0} \\ \color{green}(+3.1)}
                   & \breakInTable{\textbf{55.0} \\ \color{green}(+2.5)} 
                   & \breakInTable{\textbf{80.8} \\ \color{green}(+0.3)}\\
          \bottomrule
        \end{tabular}
    }
\end{table*}

\begin{figure*}[t]
    \centering
    \small
    \tabcolsep=0.3mm
    \begin{tabular}{c|ccc}
    \toprule
      \textbf{Method}  & \textbf{THumB} & \textbf{Flickr8k-Expert} & \textbf{Composite} \\
    \midrule
        Vanilla &
      \begin{minipage}[c]{0.30\linewidth}
        \centering
        \includegraphics[width=\linewidth]{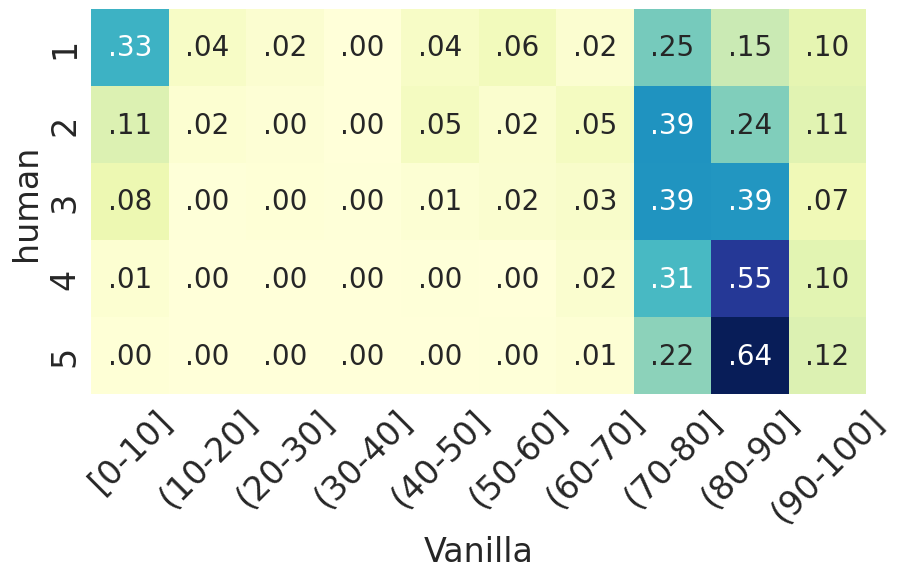}
      \end{minipage} &
      \begin{minipage}[c]{0.30\linewidth}
        \centering
        \includegraphics[width=\linewidth]{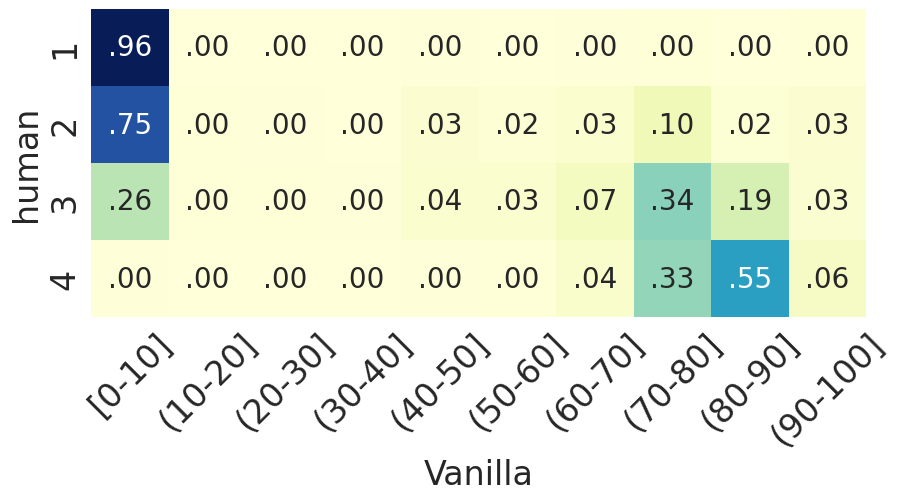}
      \end{minipage} &
      \begin{minipage}[c]{0.30\linewidth}
        \centering
        \includegraphics[width=\linewidth]{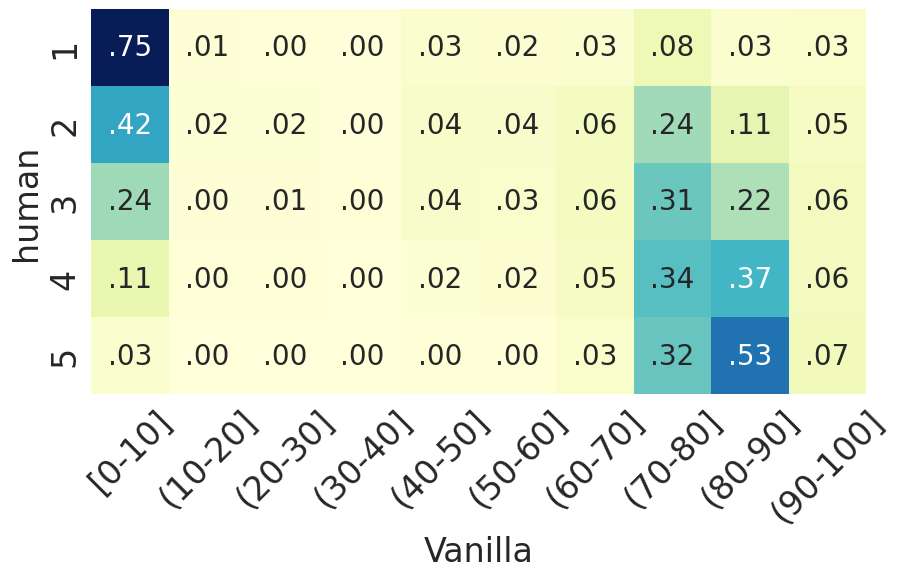}
      \end{minipage} \\

      \midrule
      \textbf{\methodName} & 
      \begin{minipage}[c]{0.30\linewidth}
        \centering
        \includegraphics[width=\linewidth]{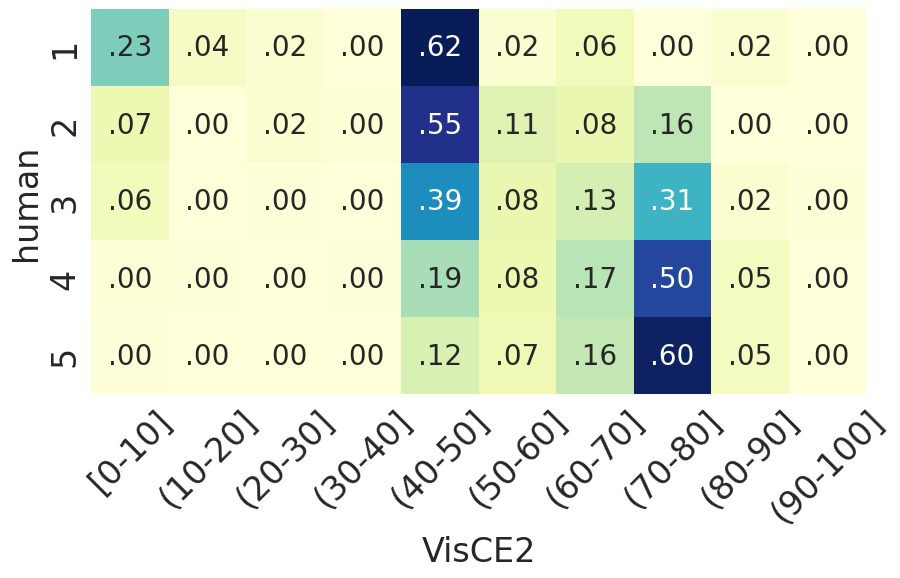}
      \end{minipage} &
      \begin{minipage}[c]{0.30\linewidth}
        \centering
        \includegraphics[width=\linewidth]{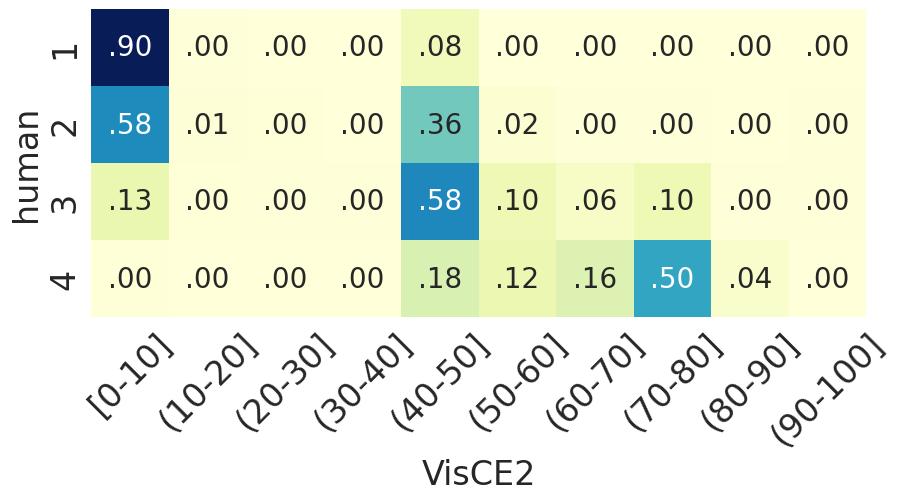}
      \end{minipage} &
      \begin{minipage}[c]{0.30\linewidth}
        \centering
        \includegraphics[width=\linewidth]{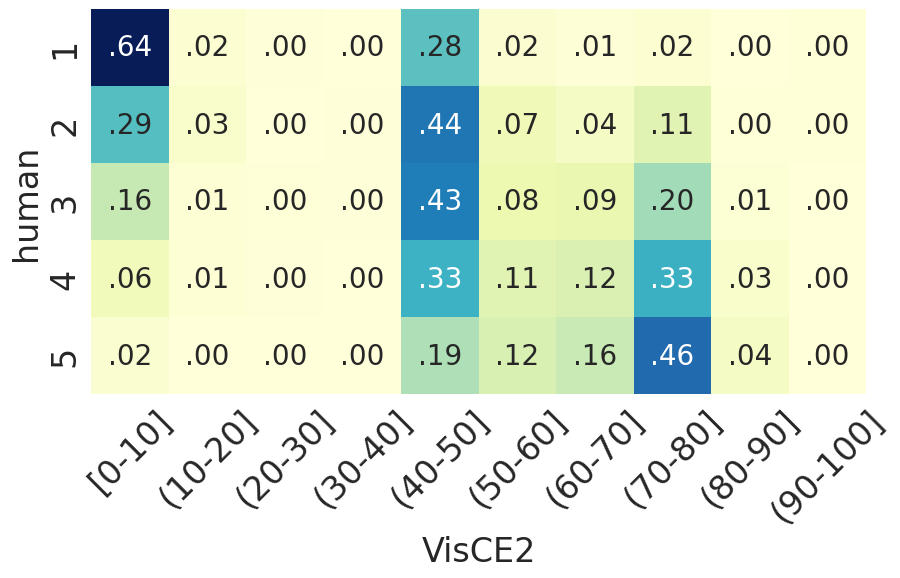}
      \end{minipage} \\
    \bottomrule
    \end{tabular}
    \caption{
         Heatmaps of human rating and automatic evaluation scores on THumB (left), Flickr8k-Expert (mid) and Composite (right). Normalized for each human evaluation score (i.e., rows). The human evaluation of THumB is referenced to the total score. 
    }
    \label{fig:fig2-score-distribution}
\end{figure*}

\subsection{Datasets}
We conduct a meta-evaluation of automatic evaluation metrics across four image captioning datasets: THumB 1.0~\cite{THumB}, Flickr8k-Expert~\cite{Flickr8k}, Composite~\cite{Composite}, and Pascal-50S~\cite{CIDEr}.

\paragraph{THumB 1.0\rm~\cite{THumB}} consists of 500 images sourced from MSCOCO, each assigned one human-written caption and four automatic-generated captions. 
The evaluation of candidate captions involves manual assessment based on three criteria: precision (how precisely the caption describes the image), recall (how well the caption covers the salient information in the image), and total (the overall quality of the caption, including fluency, inclusive language, and conciseness).

\paragraph{Flickr8k-Expert\rm~\cite{Flickr8k}} contains 5,644 pairs of images collected from Flickr and automatically generated captions, each evaluated by three experts. A score of 1 indicates that the caption is unrelated to the image, and a score of 4 indicates that it accurately describes the image. 

\paragraph{Composite\rm~\cite{Composite}} includes 2,007 images sourced from MSCOCO~\cite{MSCOCO}, 997 images from Flickr8k~\cite{Flickr8k}, and 991 images from Flickr30K~\cite{Flickr30k}. 
Each image is assigned two automatically generated captions and one human-written caption. 
Candidate captions are evaluated on a scale from 1 (irrelevant) to 5 (ideally related).

\paragraph{Pascal-50S\rm~\cite{CIDEr}} comprises 4,000 images from the UIUC Pascal sentence dataset~\cite{UIUC-Pascal}, each paired with candidate captions: one written by a human and the others automatically generated using five different methods.
Annotators were asked to determine which caption was more similar to the reference.

\subsection{Effectiveness of Visual Context}

We initially investigated the impact of incorporating visual context into the auto-evaluation method.
In addition to the candidate captions and images as VLM input, we compare the following four types of text: (i) \emph{Vanilla} uses only the task instruction,
(ii) \emph{Reference} employs references provided within datasets,
(iii) \emph{Description} incorporates detailed captions generated by LLaVA-1.5,
(iv) \emph{\methodName} (\emph{Ours}) leverages visual contexts extracted during the initial step of \methodName. See Appendix~\ref{sec:app-prompt} for the prompts in each setting.

Table \ref{tab:tab1-effect-visual-context} presents the results for different textual inputs of VLM. Our \methodName demonstrated superior evaluation performance across all datasets, suggesting that incorporating visual context enhances overall evaluation performance, except recall. In contrast, other methods, such as \emph{Reference} and \emph{Description}, showed no improvement or even degradation from the \emph{Vanilla} on some datasets. This suggests that the visual context's structured and comprehensive nature enables consistent caption evaluations. 

To conduct further analysis on \methodName, we compared the distribution of scores assigned to captions on three different datasets in the Vanilla and \methodName settings. 
The heatmap presented in Figure \ref{fig:fig2-score-distribution} plots the distribution of human and automatic evaluation scores in THumB, Flickr8k-Expert, and Composite, which comprise Likert scale human evaluations. Heatmaps are normalized by human rating points to remove sample size bias in the data set. Thus, each row sums to 1, and the heatmap rows represent the ratio of the automatic rating score to a caption of a specific human rating.
We observed that the score distribution in the Vanilla setting has two prominent peaks, with captions classified as good (70+) and bad (0-10), whereas \methodName exhibits three distant peaks: good (70-80), bad (0-10), and fair (40-50).
Furthermore, in THumB and Composite, the \methodName heatmap shows a decrease in the value of the upper right corner when compared to the Vanilla heatmap, indicating a reduction in the percentage of overrated captions.
These distribution changes suggest that introducing visual context contributes to a closer evaluation of the human impression. 

\begin{table}[t]
    \centering
    \small
     \caption{
      Correlation (Pearson's $\rho$) between baseline metrics and human judgment on THumB 1.0. ``w/o'' means discarding human annotated samples. \textbf{Bold} fonts for best score among reference- and tune- free models. $\dagger$: The scores reported in previous works. 
    }
    \label{tab:tab2-thumb}
    \scalebox{1}{
            \tabcolsep = 0.8mm
            \begin{tabular}{lcc|ccc|ccc}
              \toprule
                 \multirow{2}{*}{\textbf{Method}} & 
                 \multirow{2}{*}{\breakInTable{\textbf{Ref-}\\\textbf{free}}} & 
                 \multirow{2}{*}{\breakInTable{\textbf{Tune-}\\\textbf{free}}} &
                 \multicolumn{3}{c}{\textbf{THumB w/o}} & \multicolumn{3}{c}{\textbf{THumB w/ }} \\
                 & & & P & R & Total  & P & R & Total \\
              \midrule
                {BLEU-4}      & {\yes} & {\no} & {.21} & {.13} & {.25} & {.15} & {.04} & {.13} \\
                {ROUGE}       & {\yes} & {\no} & {.26} & {.17} & {.31} & {.18} & {.07} & {.18} \\
                {CIDEr}       & {\yes} & {\no} & {.27} & {.18} & {.33} & {.21} & {.11} & {.23} \\
                {SPICE}       & {\yes} & {\no} & {.26} & {.15} & {.30} & {.20} & {.09} & {.21} \\
                {RefCLIP-S}   & {\yes} & {\no} & {.34} & {.27} & {.44} & {.31} & {.26} & {.41} \\
      $^\dagger${InfoMetIC}   & {\no} & {\yes} & {.22} & {.30} & {.37} & {.21} & {.32} & {.38} \\
              \midrule
                CLIP-S      & \no & \no & .18 & \textbf{.27} & .32 & .17 & \textbf{.28} & .32 \\
                \rcOurs
                \textbf{\methodName} & \no & \no & \textbf{.54} & .08 & \textbf{.45}  & \textbf{.49} & .06 & \textbf{.39} \\

            \bottomrule
            \end{tabular}
        }
\end{table}

\subsection{Correlation with Human Judgment}
\label{subsec:correlation-with-human-judgment}
We analyze the proposed \methodName metrics on the THumB1.0~\cite{THumB} dataset by assessing the correlation between automatic evaluation scores and human ratings in three aspects: precision, recall, and the total score.
Following previous studies, we used Pearson's correlation coefficient as the meta-evaluation index.

The results presented in Table~\ref{tab:tab2-thumb} highlight \methodName's outstanding performance in terms of correlation with precision, surpassing all other metrics.
It exceeded RefCLIP-S, the previous state-of-the-art metric, by 0.20 and 0.18 points in settings with and without human-written captions, respectively. This suggests that \methodName accurately evaluated precise captions.
On the other hand, \methodName's performance in recall presented a contrast, exhibiting minimal correlation.
In the THumB dataset, recall scores reflect the extent to which the caption encompasses the salient information in the image.
Meanwhile, the evaluation scores of our method were calculated by considering all objects, attributes, and their relationships within the image.
The difference between focusing on the salient objects and focusing on all objects may have led to the low correlation with recall scores.
Despite its imbalanced nature, our method is highly correlated with the total scores, indicating the overall quality of captions. 

\begin{table}[t!]
    \centering
    \small
    \caption{
      Correlation between human judgement and metrics on Flickr8k-Expert and Composite. 
    }
    \label{tab:tab3-f8kexp-cmp}
    \scalebox{0.98}{
        \tabcolsep = 0.5mm
        \begin{tabular}{lcc|cc}
          \toprule
              \textbf{Method} & 
              \breakInTable{\textbf{Ref-}\\\textbf{free}} & 
              \breakInTable{\textbf{Tune-}\\\textbf{free}} &
              \breakInTable{\textbf{Flickr8k-Exp.}  \\ Kendall's $\tau$} &
              \breakInTable{\textbf{Composite} \\ Kendall's $\tau$} \\
          \midrule
                        {BLEU-4}          & {\yes} & {\no}  & {30.6} & {28.3}\\
                        {ROUGE}           & {\yes} & {\no}  & {32.1} & {30.0} \\
                        {METEOR}          & {\yes} & {\no}  & {41.5} & {36.0} \\
                        {CIDEr}           & {\yes} & {\no}  & {43.6} & {34.9} \\
                        {SPICE}           & {\yes} & {\no}  & {51.7} & {38.8} \\
              $^\dagger${BERTScore++}     & {\yes} & {\yes}  & {48.1} & {42.3} \\
                        {RefCLIP-S}       & {\yes} & {\no}  & {52.6} & {51.2} \\
              $^\dagger${RefPAC-S}        & {\yes} & {\no}  & {55.5} & {51.5} \\
  $^\dagger${CLAIR$_{\mathrm{Claude}}$}   & \yes & \no & 56.2 & 54.2 \\
  $^\dagger${CLAIR$_{\mathrm{E}}$}        & \yes & \no & 62.7 & 59.2 \\
            $^\dagger${InfoMetIC} & \no & \yes & 54.2  & 59.2 \\
          \midrule
             CLIP-S & \no & \no & 51.1  & 49.8 \\
            $^\dagger$PAC-S  & \no & {\no} & 53.9 & .51.5 \\            
            \rcOurs
            \textbf{\methodName} & \no & {\no} & \textbf{59.0} & \textbf{55.0} \\          
          \bottomrule
        \end{tabular}
    }

\end{table}
\begin{table}[t]
    \centering
    \small
     \caption{
     Comparison VLM-based caption evaluation methods on correlation (Pearson's $\rho$) with human judgement on THumB 1.0.
    }
    \label{tab:tab4-gpt4v}
    \scalebox{0.91}{
            \tabcolsep = .3mm
            \renewcommand{\arraystretch}{1.25}
            \begin{tabular}{llccc|ccc}
              \toprule
                 \multirow{2}{*}{\textbf{VLM}} & \multirow{2}{*}{\textbf{Method}} & \multicolumn{3}{c}{\breakInTable{\textbf{THumB w/o} \\ Pearson's $\rho$}} & \multicolumn{3}{c}{\breakInTable{\textbf{THumB w/} \\ Pearson's $\rho$}} \\
                 & & P & R & Total  & P & R & Total \\
              \midrule
                 LLaVA-v1.5  & Vanilla & .44 & \textbf{.08} & .38 & .41 & \textbf{.08} & .34 \\
                 \rcOurs
                 LLaVA-v1.5 & \textbf{\methodName }
                   & \breakInTable{.54 \\ \color{green}(+.10)} 
                   & \breakInTable{\textbf{.08} \\ \color{gray}($\pm$0)} 
                   & \breakInTable{.45 \\ \color{green}(+.07)}
                   & \breakInTable{.49 \\ \color{green}(+.08)}
                   & \breakInTable{.07 \\ \color{gray}(-.01)}
                   & \breakInTable{.39 \\ \color{green}(+.05)}  \\
              \midrule
                GPT-4V  & Vanilla & .53 & .03 & .41 & .50 & .05 & .38 \\
        \rcOurs GPT-4V  & \textbf{\methodName  }
                   & \breakInTable{\textbf{.58} \\ \color{green}(+.05)} 
                   & \breakInTable{.06 \\ \color{green}(+.03)} 
                   & \breakInTable{\textbf{.46} \\ \color{green}(+.05)}
                   & \breakInTable{\textbf{.55} \\ \color{green}(+.05)}
                   & \breakInTable{\textbf{.08} \\ \color{green}(+.03)}
                   & \breakInTable{\textbf{.44} \\ \color{green}(+.06)}  \\
            \bottomrule
            \end{tabular}
        }
\end{table}

\begin{figure*}[t]
    \centering
    \footnotesize
    \tabcolsep=1mm
    \scalebox{1.05}{
        \begin{tabular}{clccc}
            \toprule
                \textbf{Image} & \textbf{Candidate Caption} & Human & \textbf{\methodName} & CLIP-S \\
            \midrule
                \raisebox{1pt}{
                \multirow{3}{*}{
                    \centering
                    \includegraphics[height=16mm]{figures/qual_1_COCO_val2014_000000408774.jpg}
                }}   & a group of people on a field playing baseball. & 5 & 85 & 0.31 \rule[-6pt]{0pt}{12pt}\\
                    & a baseball player swinging a bat at a ball . & 1 &  0 & 0.26 \rule[-6pt]{0pt}{12pt}\\
                    & \begin{tabular}{l} 
                    The scene contains people wear hats and greenery and \\ people wear helmets and people wear sports dresses and crowd. \end{tabular} & 3 & 50 & 0.20 \\

            \midrule
                \raisebox{1pt}{
                \multirow{3}{*}{
                    \centering
                    \includegraphics[height=17mm]{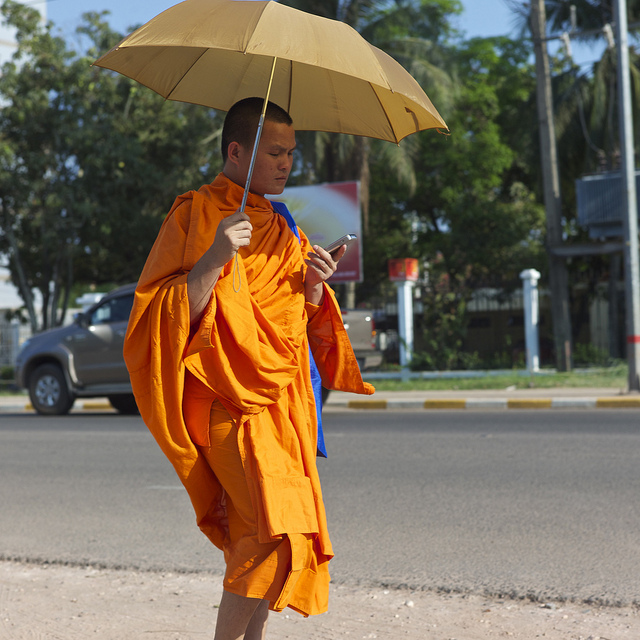}
                }}   & a man in orange garb carrying a umbrella and cell phone. & 5 & 80 & 0.41 \rule[-8pt]{0pt}{16pt}\\
                    & a woman holding an umbrella in the rain . & 1 &  0 & 0.31 \rule[-6pt]{0pt}{12pt}\\
                    & \begin{tabular}{l} 
                    The scene contains street and people walk \\
                    and booths and cars arranged in some fashion and cars. \end{tabular} & 2 & 50 & 0.23 \\

            \midrule
                \raisebox{6pt}{
                \multirow{3}{*}{
                    \centering
                    \includegraphics[height=15mm]{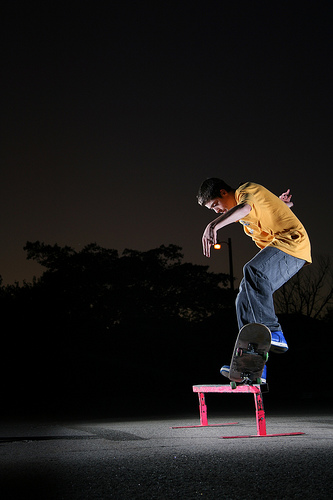}
                } }  & 
                        the boy in the yellow shirt is riding
                        a skateboard along a red bar.
                       & 5 & 85 & 0.28 \rule[-12pt]{0pt}{24pt}\\
                    &  
                        a man in a red shirt and black pants  is jumping on a skateboard . 
                       & 3 &  50 & 0.33 \rule[-9pt]{0pt}{18pt}\\
            \bottomrule
        \end{tabular}
    }
    
    \caption{
         Comparison between evaluation scores of \methodName, that of CLIP-S, and human ratings for candidate caption for images from the Composite dataset.
    }
    \label{fig:fig3-qualitative}
\end{figure*}

We also test the caption evaluation capability of \methodName using Flickr8k-Expert~\cite{Flickr8k} and Composite~\cite{Composite} dataset.
Kendall's rank correlation coefficient was used as a meta-evaluation index to measure the correlation between automatic and human evaluation.

\methodName performed at a high level compared to the other automatic evaluation metrics in its correlation with human evaluation (Table \ref{tab:tab3-f8kexp-cmp}). \methodName in the Flickr8k-Expert achieved state-of-the-art performance by a significant margin of more than 4.8 pts compared to other reference-free metrics, demonstrating that our method is highly capable of estimating the quality of image captions aligned with human judgments. 
A similar trend was observed in the Composite. Compared to CLIP-S and PAC-S, \methodName surpassed them by over 3pts. It was found that our method fell short of InfoMetIC's performance. This degradation is assumed to originate from the training method of InfoMetIC, a fine-tuning quality estimator on the three datasets that comprise the Composite dataset. 
The result also offers insight into the quality estimation based on large-scale models.
While CLAIR$_\mathrm{E}$, which utilizes an ensemble of non-public LLMs such as ChatGPT, PaLM, and Claude, achieved significant results in the benchmarks, \methodName outperformed the outcome of CLAIR$_\mathrm{Claude}$, which solely exploits Claude. This distinction highlights the efficacy of our method, considering its reliance on publicly available and comparatively smaller models. Such performance showcases the potential of more accessible models to achieve high-quality benchmarks.

\section{Analysis}

\subsection{Evaluation Performance of GPT-4V}
\label{subsec:gpt4v}
We explore the proposed method's current upper bounds and effectiveness using GPT-4V. 
GPT-4 exhibits superior performance on many vision and language tasks~\cite{Dawn-GPT4V}, and it has also been quantitatively validated to outperform LLaVA-v1.5 in several tasks~\cite{Eval-GPT4V-LLaVA15-VQA,Eval-GPT4V-LLaVA15-Bias}.
Due to budgetary constraints, we selected the relatively small yet finely-annotated dataset, THumB, used in Section~\ref{subsec:correlation-with-human-judgment}, to compare the open-source VLM model, LLaVA-v1.5, with GPT-4V using the different prompt settings (Vanilla and \methodName).
We used the \texttt{gpt-4-vision-preview} as the GPT-4V model through the Azure OpenAI API.

As shown in Table~\ref{tab:tab4-gpt4v}, the performance of GPT-4V outperformed that of LLaVA-v1.5 for precision and total score at each prompt setting.
Similar to the LLaVA-v1.5 model, applying \methodName to GPT-4 enhances the evaluation performance, indicating that the proposed method is effective even with more advanced VLMs.
In addition, GPT-4V with \methodName achieved state-of-the-art performance in terms of correlations with precision and total score.
Furthermore, the evaluation performance of the LLaVA-v1.5 model with \methodName applied was improved to the same extent as the performance of GPT-4V in the Vanilla setting. 
The effectiveness of our auto-evaluation method using this open-source VLM suggests that it can contribute to reducing evaluation costs.

\begin{figure}[t]
    \centering
    \footnotesize
    \tabcolsep=0.7mm
    \scalebox{1}{
        \begin{tabular}{cccccc}
            \toprule
                \multicolumn{6}{c}{
                    \centering
                    \includegraphics[width=.88\linewidth]{figures/COCO_val2014_000000521509.jpg}
                } \\
                  \multirow{5}{*}{\textbf{Refs.}}  
                    & \multicolumn{5}{c}{\begin{tabularx}{.80\linewidth}{X}``a lady in a bedroom with a tripod.''\end{tabularx}} \\
                    & \multicolumn{5}{c}{\begin{tabularx}{.80\linewidth}{X}``a girl sitting on a bed is taking a self portrait.''\end{tabularx}} \\
                    & \multicolumn{5}{c}{\begin{tabularx}{.80\linewidth}{X}``a girl sits in front of a mirror with a camera''\end{tabularx}} \\
                    & \multicolumn{5}{c}{
                        \begin{tabularx}{.80\linewidth}{X}
                            ``girl is setting up a camera to film herself in bed.''
                        \end{tabularx}
                    } \\
                    & \multicolumn{5}{c}{
                        \begin{tabularx}{.80\linewidth}{X}
                            ``a girl sits on the bed and photographs herself in the mirror with her tripod.''
                        \end{tabularx}
                    } \\
            
            \midrule    
                \textbf{Cand. 1} & \multicolumn{5}{c}{
                \begin{tabularx}{.80\linewidth}{X}
                     The image depicts a room that appears to be a bedroom. In the center of the image is a bed with a wooden headboard, and there's a person sitting on the bed with their legs crossed. The person is facing the camera, which is mounted on a tripod positioned in the foreground of the image. To the left and right of the bed are matching nightstands with lamps. Above the bed is a framed picture hanging on the wall, and to the right, there's a partially drawn curtain that allows daylight to enter the room. The room has a cozy, lived-in feel with various personal items scattered around, and the lighting gives the photograph a warm, muted tone.
                \end{tabularx}
                } \\
            \midrule
                BLEU-4 & METEOR & ROUGE & CIDEr & SPICE & \textbf{\methodName} \\
                0.0000 & 0.1416 & 0.1280 & 0.0000 & 0.1081 & 85 \\
            \midrule    
                \textbf{Cand. 2} & \multicolumn{5}{c}{
                \begin{tabularx}{.80\linewidth}{X}
                    The image shows a bedroom with two nightstands on either side of a bed, each with a lamp.
                \end{tabularx}
                } \\
            \midrule
                BLEU-4 & METEOR & ROUGE & CIDEr & SPICE & \textbf{\methodName} \\
                0.0000 & 0.1974 & 0.4150 & 0.0000 & 0.1212 & 80 \\
            \bottomrule
        \end{tabular}
    }
    
    \caption{
         Comparison between evaluation scores of \methodName and classic reference-based metrics. The image and the references are from the MS-COCO dataset. Both captions are automatically generated by GPT-4V.
    }
    \label{fig:fig4-compare-refbased}
\end{figure}

\subsection{Qualitative Analysis}
\label{subsec:qualitative}
To clarify the differences in trends among other reference-free metrics, we compared the evaluation scores assigned to the images by \methodName with those of CLIP-S, a typical reference-free metric.
Figure \ref{fig:fig3-qualitative} presents examples of human ratings and scores of the auto-eval methods for image-caption pairs. 
These qualitative examples indicate the discriminative ability of \methodName to the accuracy of captions. 
CLIP-S tends to overestimate captions describing the presence of objects in the image.
For example, CLIP-S assigns a relatively high score to the incorrect caption \emph{``a baseball player swinging a bat at a ball.''} to the image in the first row.
However, although ``the baseball player'' is present in the image, the event of ``swinging the bat at the ball'' has not occurred.
Humans can detect such contradiction accurately, whereas CLIP-S often ignores it.
In the second example, CLIP-S also overestimated incorrect image descriptions containing the salient object, an umbrella, in the image.
A previous study~\cite{ahmadi2023examination} has also pointed out that CLIP-S tends to be affected by the presence or absence of descriptions of salient objects in the image, confirmed in these examples.
The third example shows that CLIP-S overlooks mistakes in combining objects and their colors.
In contrast, \methodName can identify the discrepancy using the visual context and accurately assess image-caption pairs.

\subsection{Comparison to Reference-based Metrics}
We present another qualitative example to shed light on the effectiveness of our proposed method as well as the limitations of existing reference-based methods. 
Figure~\ref{fig:fig4-compare-refbased} provides an image, five annotated reference captions drawn from the MS-COCO dataset, and two generated caption candidates with the recent image captioning model of GPT-4V with automatic evaluation scores.
The first candidate is a caption generated using the straightforward prompt, \emph{``Generate a detailed description for the given image.''} This caption accurately describes the image's content in quite a detailed manner.~\footnote{CLIP-S and RefCLIP-S cannot handle such extended captions correctly due to the context length.} 
Indeed, its detailedness overwhelms those of five human-written references, and hence, all of the reference-based metrics undervalued the caption.
The other candidate's caption follows the style of references, providing accurate descriptions of some objects, although it falls short in describing overall details.
This candidate caption, however, achieves higher metrics in some auto-evaluated scores.
We assume this is because the sophisticatedness of the recent image captioning models exceeds those of the reference-based metrics, and hence reference-free metrics, such as \methodName, have a clear merit for evaluating current overwhelmingly detailed image captions.

\section{Conclusion}
We have proposed \methodName, a prompting method for automatic VLM-based evaluation of image captions.
\methodName deviates from the traditional method of evaluating similarity using only reference text and images, weaving visual context into the evaluation framework with a compositional form. This technique allows the VLMs to understand the detailed visual dependencies of images better and validate them based on more exhaustive content than those that the reference captions provide. Meta-evaluation experiments revealed that scores output by \methodName have excellent consistency with human judgments, especially in caption accuracy, which outperforms existing evaluation metrics. 
Future work includes providing VLM-based automatic evaluation along various perspectives, which would provide even more fine-grained information for humans.

\section*{Limitations}

\paragraph{Evaluation cost}
Our proposed method, which utilizes VLMs for evaluation, requires a higher computational cost than previous approaches. In our experiments, we employed an NVIDIA A100 40GB GPU, and extracting visual context took about 10 seconds per sample and caption evaluation 100 milliseconds, respectively.
This limitation significantly depends on the inference speed of VLMs, suggesting that future improvements in VLMs could mitigate this issue.

\paragraph{Sensitivity to the prompt}
~\citet{ButterFlyEffect-of-Prompt} points out that Large Language Models (LLMs) performance varies based on the provided prompts.
Although this is an issue for LLMs, it is natural to consider it for VLMs, even if the impact is not quantified yet. Understanding how VLMs respond to different prompts is key to making them more reliable, which remains for future work.

\section*{Ethics Statement}
As our proposed method \methodName is utilized for providing scores to captions, we do not anticipate the negative impacts of this work. However, as with other machine learning methods, we recommend exercising caution.

\section*{Acknowledgements}
This work was supported by JSPS KAKENHI Grant Number JP22K17983, and by JST PRESTO Grant Number JPMJPR20C2.

\bibliography{custom}

\begin{thebibliography}{53}
\expandafter\ifx\csname natexlab\endcsname\relax\def\natexlab#1{#1}\fi

\bibitem[{Aditya et~al.(2015)Aditya, Yang, Baral, Fermuller, and Aloimonos}]{Composite}
Somak Aditya, Yezhou Yang, Chitta Baral, Cornelia Fermuller, and Yiannis Aloimonos. 2015.
\newblock From images to sentences through scene description graphs using commonsense reasoning and knowledge.
\newblock \emph{arXiv preprint arXiv:1511.03292}.

\bibitem[{Ahmadi and Agrawal(2023)}]{ahmadi2023examination}
Saba Ahmadi and Aishwarya Agrawal. 2023.
\newblock \href {http://arxiv.org/abs/2305.14998} {An examination of the robustness of reference-free image captioning evaluation metrics}.
\newblock \emph{arXiv preprint arXiv:2305.14998}.

\bibitem[{Anderson et~al.(2016)Anderson, Fernando, Johnson, and Gould}]{SPICE}
Peter Anderson, Basura Fernando, Mark Johnson, and Stephen Gould. 2016.
\newblock {SPICE:} semantic propositional image caption evaluation.
\newblock In \emph{ECCV}, pages 382--398.

\bibitem[{Bai et~al.(2023)}]{Qwen-VL}
Jinze Bai et~al. 2023.
\newblock {Q}wen-{VL}: A versatile vision-language model for understanding, localization, text reading, and beyond.
\newblock \emph{arXiv preprint arXiv:2308.12966}.

\bibitem[{Bai et~al.(2022)}]{Claude}
Yuntao Bai et~al. 2022.
\newblock \href {http://arxiv.org/abs/2204.05862} {Training a helpful and harmless assistant with reinforcement learning from human feedback}.
\newblock \emph{arXiv preprint arXiv:2204.05862}.

\bibitem[{Chan et~al.(2023)Chan, Petryk, Gonzalez, Darrell, and Canny}]{CLAIR}
David~M Chan, Suzanne Petryk, Joseph~E Gonzalez, Trevor Darrell, and John Canny. 2023.
\newblock {CLAIR: Evaluating Image Captions with Large Language Models}.
\newblock In \emph{EMNLP}, Singapore, Singapore. Association for Computational Linguistics.

\bibitem[{Chen et~al.(2023)Chen, Zhu, Shen, Li, Liu, Zhang, Krishnamoorthi, Chandra, Xiong, and Elhoseiny}]{MiniGPT-v2}
Jun Chen, Deyao Zhu, Xiaoqian Shen, Xiang Li, Zechu Liu, Pengchuan Zhang, Raghuraman Krishnamoorthi, Vikas Chandra, Yunyang Xiong, and Mohamed Elhoseiny. 2023.
\newblock Minigpt-v2: large language model as a unified interface for vision-language multi-task learning.
\newblock \emph{arXiv preprint arXiv:2310.09478}.

\bibitem[{Chen et~al.(2015)}]{MSCOCO}
Xinlei Chen et~al. 2015.
\newblock Microsoft {COCO} captions: Data collection and evaluation server.
\newblock \emph{arXiv:1504.00325}.

\bibitem[{Chen et~al.(2020)Chen, Li, Yu, Kholy, Ahmed, Gan, Cheng, and Liu}]{UNITER}
Yen-Chun Chen, Linjie Li, Licheng Yu, Ahmed~El Kholy, Faisal Ahmed, Zhe Gan, Yu~Cheng, and Jingjing Liu. 2020.
\newblock Uniter: Universal image-text representation learning.
\newblock In \emph{ECCV}.

\bibitem[{Chowdhery et~al.(2022)}]{PaLM}
Aakanksha Chowdhery et~al. 2022.
\newblock \href {http://arxiv.org/abs/2204.02311} {{P}a{LM}: Scaling language modeling with pathways}.
\newblock \emph{arXiv preprint arXiv:2204.02311}.

\bibitem[{Cui et~al.(2023)Cui, Zhou, Yang, Wu, Zhang, Zou, and Yao}]{Eval-GPT4V-LLaVA15-Bias}
Chenhang Cui, Yiyang Zhou, Xinyu Yang, Shirley Wu, Linjun Zhang, James Zou, and Huaxiu Yao. 2023.
\newblock \href {http://arxiv.org/abs/2311.03287} {Holistic analysis of hallucination in gpt-4v(ision): Bias and interference challenges}.

\bibitem[{Dai et~al.(2023)}]{InstructBLIP}
Wenliang Dai et~al. 2023.
\newblock {I}nstruct{BLIP}: Towards general-purpose vision-language models with instruction tuning.
\newblock \emph{arXiv preprint arXiv:2305.06500}.

\bibitem[{Denkowski and Lavie(2014)}]{METEOR}
Michael Denkowski and Alon Lavie. 2014.
\newblock \href {https://doi.org/10.3115/v1/W14-3348} {Meteor universal: Language specific translation evaluation for any target language}.
\newblock In \emph{WMT}, pages 376--380.

\bibitem[{Devlin et~al.(2018)Devlin, Chang, Lee, and Toutanova}]{BERT}
Jacob Devlin, Ming-Wei Chang, Kenton Lee, and Kristina Toutanova. 2018.
\newblock {BERT: Pre-training of Deep Bidirectional Transformers for Language Understanding}.
\newblock \emph{arXiv preprint arXiv:1810.04805}.

\bibitem[{Google(2023)}]{Gemini}
Google. 2023.
\newblock \href {http://arxiv.org/abs/2312.11805} {Gemini: A family of highly capable multimodal models}.

\bibitem[{Hessel et~al.(2021)Hessel, Holtzman, Forbes, Bras, and Choi}]{CLIPScore}
Jack Hessel, Ari Holtzman, Maxwell Forbes, Ronan~Le Bras, and Yejin Choi. 2021.
\newblock {CLIPScore:} a reference-free evaluation metric for image captioning.
\newblock In \emph{EMNLP}.

\bibitem[{Hodosh et~al.(2013)Hodosh, Young, and Hockenmaier}]{Flickr8k}
Micah Hodosh, Peter Young, and Julia Hockenmaier. 2013.
\newblock \href {http://dblp.uni-trier.de/db/journals/jair/jair47.html#HodoshYH13} {Framing image description as a ranking task: Data, models and evaluation metrics.}
\newblock \emph{J. Artif. Intell. Res.}, 47:853--899.

\bibitem[{Hu et~al.(2023)Hu, Chen, Zhang, and Jin}]{InfoMetIC}
Anwen Hu, Shizhe Chen, Liang Zhang, and Qin Jin. 2023.
\newblock {InfoMetIC: An Informative Metric for Reference-free Image Caption Evaluation}.
\newblock In \emph{{ACL}}, pages 3171--3185. Association for Computational Linguistics.

\bibitem[{Jiang et~al.(2019)Jiang, Huang, Zhang, Wang, Zhang, Gan, Diesner, and Gao}]{TIGEr}
Ming Jiang, Qiuyuan Huang, Lei Zhang, Xin Wang, Pengchuan Zhang, Zhe Gan, Jana Diesner, and Jianfeng Gao. 2019.
\newblock \href {https://doi.org/10.18653/v1/D19-1220} {{TIGE}r: Text-to-image grounding for image caption evaluation}.
\newblock In \emph{Proceedings of the 2019 Conference on Empirical Methods in Natural Language Processing and the 9th International Joint Conference on Natural Language Processing (EMNLP-IJCNLP)}, pages 2141--2152, Hong Kong, China. Association for Computational Linguistics.

\bibitem[{Kasai et~al.(2022)Kasai, Sakaguchi, Dunagan, Morrison, Le~Bras, Choi, and Smith}]{THumB}
Jungo Kasai, Keisuke Sakaguchi, Lavinia Dunagan, Jacob Morrison, Ronan Le~Bras, Yejin Choi, and Noah~A. Smith. 2022.
\newblock \href {https://doi.org/10.18653/v1/2022.naacl-main.254} {Transparent human evaluation for image captioning}.
\newblock In \emph{NAACL}, pages 3464--3478, Seattle, United States. Association for Computational Linguistics.

\bibitem[{Kulkarni et~al.(2011)Kulkarni, Premraj, Dhar, Li, Choi, Berg, and Berg}]{BabyTalk}
Girish Kulkarni, Visruth Premraj, Sagnik Dhar, Siming Li, Yejin Choi, Alexander~C Berg, and Tamara~L Berg. 2011.
\newblock \href {https://doi.org/10.1109/CVPR.2011.5995466} {Baby talk: Understanding and generating simple image descriptions}.
\newblock In \emph{CVPR 2011}, pages 1601--1608.

\bibitem[{Lee et~al.(2021)Lee, Yoon, Dernoncourt, Bui, and Jung}]{UMIC}
Hwanhee Lee, Seunghyun Yoon, Franck Dernoncourt, Trung Bui, and Kyomin Jung. 2021.
\newblock \href {https://doi.org/10.18653/v1/2021.acl-short.29} {{{UMIC}: An Unreferenced Metric for Image Captioning via Contrastive Learning}}.
\newblock In \emph{ACL-IJCNLP}, pages 220--226, Online. Association for Computational Linguistics.

\bibitem[{Lee et~al.(2020)Lee, Yoon, Dernoncourt, Kim, Bui, and Jung}]{ViLBERTScore}
Hwanhee Lee, Seunghyun Yoon, Franck Dernoncourt, Doo~Soon Kim, Trung Bui, and Kyomin Jung. 2020.
\newblock \href {https://doi.org/10.18653/v1/2020.eval4nlp-1.4} {{V}i{LBERTS}core: Evaluating image caption using vision-and-language {BERT}}.
\newblock In \emph{Proceedings of the First Workshop on Evaluation and Comparison of NLP Systems}, pages 34--39, Online. Association for Computational Linguistics.

\bibitem[{Lee et~al.(2018)Lee, Chen, Hua, Hu, and He}]{SCAN}
Kuang-Huei Lee, Xi~Chen, Gang Hua, Houdong Hu, and Xiaodong He. 2018.
\newblock Stacked cross attention for image-text matching.
\newblock In \emph{Proceedings of the European conference on computer vision (ECCV)}, pages 201--216.

\bibitem[{Li et~al.(2023)Li, Li, Savarese, and Hoi}]{BLIP-2}
Junnan Li, Dongxu Li, Silvio Savarese, and Steven Hoi. 2023.
\newblock \href {http://arxiv.org/abs/2301.12597} {Blip-2: Bootstrapping language-image pre-training with frozen image encoders and large language models}.

\bibitem[{Li et~al.(2022)Li, Li, Xiong, and Hoi}]{BLIP}
Junnan Li, Dongxu Li, Caiming Xiong, and Steven Hoi. 2022.
\newblock \href {http://arxiv.org/abs/2201.12086} {Blip: Bootstrapping language-image pre-training for unified vision-language understanding and generation}.

\bibitem[{Li et~al.(2024)Li, Wang, Hu, Chen, Zhong, Lyu, Wang, and Zhang}]{Eval-GPT4V-LLaVA15-VQA}
Yunxin Li, Longyue Wang, Baotian Hu, Xinyu Chen, Wanqi Zhong, Chenyang Lyu, Wei Wang, and Min Zhang. 2024.
\newblock \href {http://arxiv.org/abs/2311.07536} {A comprehensive evaluation of gpt-4v on knowledge-intensive visual question answering}.

\bibitem[{Lin(2004)}]{ROUGE}
Chin-Yew Lin. 2004.
\newblock \href {https://aclanthology.org/W04-1013} {{ROUGE}: A package for automatic evaluation of summaries}.
\newblock In \emph{Text Summarization Branches Out}, pages 74--81, Barcelona, Spain. Association for Computational Linguistics.

\bibitem[{Liu et~al.(2023{\natexlab{a}})Liu, Li, Li, and Lee}]{LLaVA1.5}
Haotian Liu, Chunyuan Li, Yuheng Li, and Yong~Jae Lee. 2023{\natexlab{a}}.
\newblock Improved baselines with visual instruction tuning.
\newblock \emph{arXiv preprint arXiv:2310.03744}.

\bibitem[{Liu et~al.(2023{\natexlab{b}})Liu, Li, Wu, and Lee}]{LLaVA}
Haotian Liu, Chunyuan Li, Qingyang Wu, and Yong~Jae Lee. 2023{\natexlab{b}}.
\newblock Visual instruction tuning.
\newblock \emph{arXiv preprint arXiv:2304.08485}.

\bibitem[{Lu et~al.(2019)Lu, Batra, Parikh, and Lee}]{ViLBERT}
Jiasen Lu, Dhruv Batra, Devi Parikh, and Stefan Lee. 2019.
\newblock \href {http://arxiv.org/abs/1908.02265} {{ViLBERT: Pretraining Task-Agnostic Visiolinguistic Representations for Vision-and-Language Tasks}}.

\bibitem[{Lu et~al.(2017)Lu, Xiong, Parikh, and Socher}]{Lu-etal-2017-knowing}
Jiasen Lu, Caiming Xiong, Devi Parikh, and Richard Socher. 2017.
\newblock \href {https://doi.org/10.1109/CVPR.2017.345} {Knowing when to look: Adaptive attention via a visual sentinel for image captioning}.
\newblock In \emph{2017 IEEE Conference on Computer Vision and Pattern Recognition (CVPR)}, pages 3242--3250.

\bibitem[{Mitchell et~al.(2012)Mitchell, Dodge, Goyal, Yamaguchi, Stratos, Han, Mensch, Berg, Berg, and Daum{\'e}~III}]{Midge}
Margaret Mitchell, Jesse Dodge, Amit Goyal, Kota Yamaguchi, Karl Stratos, Xufeng Han, Alyssa Mensch, Alex Berg, Tamara Berg, and Hal Daum{\'e}~III. 2012.
\newblock \href {https://aclanthology.org/E12-1076} {{M}idge: Generating image descriptions from computer vision detections}.
\newblock In \emph{Proceedings of the 13th Conference of the {E}uropean Chapter of the Association for Computational Linguistics}, pages 747--756, Avignon, France. Association for Computational Linguistics.

\bibitem[{OpenAI(2022)}]{ChatGPT}
OpenAI. 2022.
\newblock Introducing {C}hat{GPT}.

\bibitem[{OpenAI et~al.(2023)}]{GPT-4}
OpenAI et~al. 2023.
\newblock \href {http://arxiv.org/abs/2303.08774} {Gpt-4 technical report}.
\newblock \emph{arXiv preprint arXiv:2303.08774}.

\bibitem[{Papineni et~al.(2002)Papineni, Roukos, Ward, and Zhu}]{BLEU}
Kishore Papineni, Salim Roukos, Todd Ward, and Wei-Jing Zhu. 2002.
\newblock \href {https://doi.org/10.3115/1073083.1073135} {{BLEU}: a method for automatic evaluation of machine translation}.
\newblock In \emph{ACL}, pages 311--318.

\bibitem[{Radford et~al.(2021)Radford, Kim, Hallacy, Ramesh, Goh, Agarwal, Sastry, Askell, Mishkin, Clark, Krueger, and Sutskever}]{CLIP}
Alec Radford, Jong~Wook Kim, Chris Hallacy, Aditya Ramesh, Gabriel Goh, Sandhini Agarwal, Girish Sastry, Amanda Askell, Pamela Mishkin, Jack Clark, Gretchen Krueger, and Ilya Sutskever. 2021.
\newblock \href {http://arxiv.org/abs/2103.00020} {Learning transferable visual models from natural language supervision}.
\newblock \emph{arXiv preprint arXiv:2103.00020}.

\bibitem[{Rashtchian et~al.(2010)Rashtchian, Young, Hodosh, and Hockenmaier}]{UIUC-Pascal}
Cyrus Rashtchian, Peter Young, Micah Hodosh, and Julia Hockenmaier. 2010.
\newblock \href {https://aclanthology.org/W10-0721} {Collecting image annotations using {A}mazon{'}s {M}echanical {T}urk}.
\newblock In \emph{Proceedings of the {NAACL} {HLT} 2010 Workshop on Creating Speech and Language Data with {A}mazon{'}s Mechanical Turk}, pages 139--147, Los Angeles. Association for Computational Linguistics.

\bibitem[{Rombach et~al.(2022)Rombach, Blattmann, Lorenz, Esser, and Ommer}]{StableDiffusion}
Robin Rombach, Andreas Blattmann, Dominik Lorenz, Patrick Esser, and Bj\"orn Ommer. 2022.
\newblock High-resolution image synthesis with latent diffusion models.
\newblock In \emph{Proceedings of the IEEE/CVF Conference on Computer Vision and Pattern Recognition (CVPR)}, pages 10684--10695.

\bibitem[{Salinas and Morstatter(2024)}]{ButterFlyEffect-of-Prompt}
Abel Salinas and Fred Morstatter. 2024.
\newblock \href {http://arxiv.org/abs/2401.03729} {The butterfly effect of altering prompts: How small changes and jailbreaks affect large language model performance}.

\bibitem[{Sarto et~al.(2023)Sarto, Barraco, Cornia, Baraldi, and Cucchiara}]{PAC-S}
Sara Sarto, Manuele Barraco, Marcella Cornia, Lorenzo Baraldi, and Rita Cucchiara. 2023.
\newblock {Positive-Augmented Contrastive Learning for Image and Video Captioning Evaluation}.
\newblock In \emph{CVPR}.

\bibitem[{Vaswani et~al.(2017)Vaswani, Shazeer, Parmar, Uszkoreit, Jones, Gomez, Kaiser, and Polosukhin}]{Transformer}
Ashish Vaswani, Noam Shazeer, Niki Parmar, Jakob Uszkoreit, Llion Jones, Aidan~N Gomez, \L~ukasz Kaiser, and Illia Polosukhin. 2017.
\newblock \href {https://proceedings.neurips.cc/paper_files/paper/2017/file/3f5ee243547dee91fbd053c1c4a845aa-Paper.pdf} {Attention is all you need}.
\newblock In \emph{Advances in Neural Information Processing Systems}, volume~30. Curran Associates, Inc.

\bibitem[{Vedantam et~al.(2015)Vedantam, Lawrence~Zitnick, and Parikh}]{CIDEr}
Ramakrishna Vedantam, C~Lawrence~Zitnick, and Devi Parikh. 2015.
\newblock Cider: Consensus-based image description evaluation.
\newblock In \emph{CVPR}, pages 4566--4575.

\bibitem[{Vinyals et~al.(2015)Vinyals, Toshev, Bengio, and Erhan}]{ShowAndTell}
Oriol Vinyals, Alexander Toshev, Samy Bengio, and Dumitru Erhan. 2015.
\newblock \href {https://doi.org/10.1109/CVPR.2015.7298935} {Show and tell: A neural image caption generator}.
\newblock In \emph{2015 IEEE Conference on Computer Vision and Pattern Recognition (CVPR)}, pages 3156--3164.

\bibitem[{Wang et~al.(2022)Wang, Yang, Men, Lin, Bai, Li, Ma, Zhou, Zhou, and Yang}]{OFA}
Peng Wang, An~Yang, Rui Men, Junyang Lin, Shuai Bai, Zhikang Li, Jianxin Ma, Chang Zhou, Jingren Zhou, and Hongxia Yang. 2022.
\newblock Ofa: Unifying architectures, tasks, and modalities through a simple sequence-to-sequencelearning framework.
\newblock \emph{arXiv preprint arXiv:2202.03052}.

\bibitem[{Wang et~al.(2021)Wang, Yao, Wang, Wu, and Chen}]{FAIEr}
Sijin Wang, Ziwei Yao, Ruiping Wang, Zhongqin Wu, and Xilin Chen. 2021.
\newblock {FAIEr: Fidelity and Adequacy Ensured Image Caption Evaluation}.
\newblock In \emph{Proceedings of the IEEE/CVF Conference on Computer Vision and Pattern Recognition (CVPR)}, pages 14050--14059.

\bibitem[{Xu et~al.(2017)Xu, Zhu, Choy, and Fei-Fei}]{SceneGraph}
Danfei Xu, Yuke Zhu, Christopher~B. Choy, and Li~Fei-Fei. 2017.
\newblock \href {http://arxiv.org/abs/1701.02426} {Scene graph generation by iterative message passing}.

\bibitem[{Xu et~al.(2015)Xu, Ba, Kiros, Cho, Courville, Salakhudinov, Zemel, and Bengio}]{ShowAttendTell}
Kelvin Xu, Jimmy Ba, Ryan Kiros, Kyunghyun Cho, Aaron Courville, Ruslan Salakhudinov, Rich Zemel, and Yoshua Bengio. 2015.
\newblock \href {https://proceedings.mlr.press/v37/xuc15.html} {Show, attend and tell: Neural image caption generation with visual attention}.
\newblock In \emph{Proceedings of the 32nd International Conference on Machine Learning}, volume~37 of \emph{Proceedings of Machine Learning Research}, pages 2048--2057, Lille, France. PMLR.

\bibitem[{Yang et~al.(2023)Yang, Li, Lin, Wang, Lin, Liu, and Wang}]{Dawn-GPT4V}
Zhengyuan Yang, Linjie Li, Kevin Lin, Jianfeng Wang, Chung-Ching Lin, Zicheng Liu, and Lijuan Wang. 2023.
\newblock \href {http://arxiv.org/abs/2309.17421} {The dawn of lmms: Preliminary explorations with gpt-4v(ision)}.

\bibitem[{Yi et~al.(2020)Yi, Deng, and Hu}]{BERTScorePP}
Yanzhi Yi, Hangyu Deng, and Jinglu Hu. 2020.
\newblock \href {https://doi.org/10.18653/v1/2020.acl-main.93} {{Improving Image Captioning Evaluation by Considering Inter References Variance}}.
\newblock In \emph{Proceedings of the 58th Annual Meeting of the Association for Computational Linguistics}, pages 985--994, Online. Association for Computational Linguistics.

\bibitem[{Young et~al.(2014)Young, Lai, Hodosh, and Hockenmaier}]{Flickr30k}
Peter Young, Alice Lai, Micah Hodosh, and Julia Hockenmaier. 2014.
\newblock \href {https://doi.org/10.1162/tacl_a_00166} {From image descriptions to visual denotations: New similarity metrics for semantic inference over event descriptions}.
\newblock \emph{TACL}, 2:67--78.

\bibitem[{Zhang* et~al.(2020)Zhang*, Kishore*, Wu*, Weinberger, and Artzi}]{BERTScore}
Tianyi Zhang*, Varsha Kishore*, Felix Wu*, Kilian~Q. Weinberger, and Yoav Artzi. 2020.
\newblock \href {https://openreview.net/forum?id=SkeHuCVFDr} {{BERTScore: Evaluating Text Generation with BERT}}.
\newblock In \emph{International Conference on Learning Representations}.

\bibitem[{Zhu et~al.(2023)}]{MiniGPT-4}
Deyao Zhu et~al. 2023.
\newblock {M}ini{GPT}-4: Enhancing vision-language understanding with advanced large language models.
\newblock \emph{arXiv preprint arXiv:2304.10592}.

\end{thebibliography}
\bibliographystyle{acl_natbib}

\clearpage

\onecolumn
\appendix
\section{Prompts}
\label{sec:app-prompt}
We provide the prompts used in our experiments in Table~\ref{tab:tab5-prompt}.

\begin{table*}[h]
    \centering
    \footnotesize
    \caption{
      The prompts used in the experiment. \{caption\}, \{references\}, and \{context\} indicate the place to insert. The image is given at the beginning of each prompt.
    }
    \label{tab:tab5-prompt}
    \scalebox{0.98}{
        \begin{tabularx}{\linewidth}{p{30mm}|X}
          \toprule
              \textbf{Method} & \textbf{Prompt} \\
          \midrule
         \cellcolor{purple} Vanilla & On a precise scale from 0 to 100, rate whether the candidate caption is appropriate for the given image.

Candidate caption: \{caption\} 

Your rating must be a single digit between 0 and 100.
 \\
              \midrule
         \cellcolor{purple} Reference & On a precise scale from 0 to 100, rate whether the candidate caption is appropriate for the given image.

Candidate caption: \{caption\}

Use the image and the following reference to evaluate the candidate caption:

Reference: \{references\}

Your final rating must be a single digit between 0 and 100. \\
              \midrule
         \cellcolor{yellow} Description - Step 1 &  Generate a detailed description for the given image.
 \\
              \midrule
         \cellcolor{yellow} \methodName - Step 1 & Analyze the uploaded image and provide a structured output focusing on the objects, their features, and the relationships between them. 
Select up to five of the most important elements. The output should be organized as follows: 

List of Important Objects (up to five): 

- Object 1: [Brief description]

- Object 2: [Brief description]

- (Continue as necessary, up to five objects)

Features (Specific characteristics and attributes of each object, such as color, shape, size, and texture):

- Features of Object 1: [Detailed description of features]

- Features of Object 2: [Detailed description of features]

- (Continue as necessary for each selected object)

Relationships (The way objects interact or are positioned relative to each other, without using specific object names or symbols):

- Description of a relationship: [General description]

- Another relationship: [General description]

- (Continue as necessary for each relevant relationship)

Focus on providing unique and detailed insights into the features and relationships of the selected objects up to five objects.
 \\
              \midrule
         \cellcolor{purple} 
         \begin{tabular}[t]{l}
              Description - Step 2 \\ \methodName - Step 2
         \end{tabular} &  On a precise scale from 0 to 100, rate whether the candidate caption is appropriate for the given image. 

Candidate caption: \{caption\} 

Use the image and following visual context to evaluate the candidate caption: 

Visual context: \{context\} 

Your final rating must be a single digit between 0 and 100. 
 \\
          \bottomrule
        \end{tabularx}
    }
    
\end{table*}

\begin{table}[t]
    \centering
    \footnotesize
    \caption{Accuracy on the Pascal-50S dataset. \textbf{Bold} fonts for best score among reference- and tune- free models. $\dagger$: The scores reported in previous works. }
    \label{tab:tab6-pascal50s}
    \scalebox{1}{
        \tabcolsep = 3.mm
        \begin{tabular}{lcc|ccccc}
        
            \toprule
                \multirow{2}{*}{\textbf{Method}} & \multirow{2}{*}{\textbf{Reference-free}}  & \multirow{2}{*}{\textbf{Tune-free}}  
                & \multicolumn{5}{c}{\textbf{PASCAL-50S}} \\
                                &   &   & HC & HI & HM & MM & Mean \\
            \midrule
                    {BLEU-4}      & \yes & \no & 53.0 & 92.4 & 86.7 & 59.4 & 72.8 \\
                    {ROUGE}       & \yes & \no & 51.5 & 94.5 & 92.5 & 57.7 & 74.0 \\
                    {METEOR}      & \yes & \no & 56.7 & 97.6 & 94.2 & 63.4 & 77.9 \\
                    {CIDEr}       & \yes & \no & 53.0 & 98.0 & 91.5 & 64.5 & 76.7 \\
                    {SPICE}       & \yes & \no & 52.6 & 93.9 & 83.6 & 48.1 & 69.5 \\
          $^\dagger${BERTScore++} & \yes & \yes & 65.4 & 98.1 & 96.4 & 60.3 & 80.1 \\
                    {RefCLIP-S}   & \yes & \no & 64.9 & 99.5 & 95.5 & 73.3 & 83.3 \\
          $^\dagger${RefPAC-S}    & \yes & \no & 67.7 & 99.6 & 96.0 & 75.6 & 84.7 \\
          $^\dagger${CLAIR$_{\mathrm{Claude}}$} & {\yes} & {\no} & {57.9} & {98.5} & {91.3} & {62.9} & {77.6} \\
          $^\dagger${CLAIR$_{\mathrm{E}}$} & {\yes} & {\no} & {57.7} & {99.8} & {94.6} & {75.6} & {81.9} \\
          $^\dagger${InfoMetIC}   & {\no} & {\yes} & {69.0} & {99.8} & {94.0} & {78.3} & {85.3} \\
          \midrule
                    CLIP-S      & \no & \no  & 55.9 & 99.3 & 96.5 & 72.0 & 80.9 \\
          $^\dagger$PAC-S       & \no & \no  & 60.6 & 99.3 & \textbf{96.9} & \textbf{72.9} & \textbf{82.4} \\
          \rcOurs\methodName    & \no & \no  & \textbf{60.7} & \textbf{99.6} & 93.6 & 69.3 & 80.8 \\
          \bottomrule
        \end{tabular}
    }
\end{table}

\section{Caption Pairwise Ranking}
\label{sec:app-pascal}

To further verify how accurately our \methodName determines the relative preference of captions, we conducted a meta-evaluation with the Pascal-50S dataset~\cite{CIDEr} that comprises human preference judgments indicating which of the two captions for an image is more appropriate.
The preference judgments for pairs of candidate captions are classified into the following four categories;
\begin{enumerate}
\setlength{\itemsep}{-4pt}
    \item HC: pairs of human-written captions, both of which correctly represent the image's content;
    \item HI: pairs of human-written captions where one caption is correct and the other is incorrect;
    \item HM: pairs of correct captions, one of which is human-written and the other machine-generated;
    \item MM: pairs of machine-generated captions, both of which correctly represent the image's content.
\end{enumerate}
For each category, we measured accuracy, the judgment agreement between the human preferences, and the results of the automatic evaluation method. 

The pairwise ranking agreement result shown in Table~\ref{tab:tab6-pascal50s} depicts that our method outperformed CLIP-S and PAC-S in the HC and HI categories. On the other hand, relatively low performance was observed in HM and MM compared to other metrics. This degraded performance, especially noticeable for MM, can be attributed to the nature of the candidate captions. The automatically generated captions in the Pascal-50S dataset are made with classical captioning methods, such as Midge~\cite{Midge} and Babytalk~\cite{BabyTalk}, often containing errors. Therefore, both are judged as poor-quality captions, causing the model's performance to deteriorate.
Another contributing factor might be the disagreement in objectives between human annotation and automatic quality estimation. In the annotation process for the Pascal-50S dataset, the worker selected captions that are more similar to randomly selected references, which does not necessarily equate to superiority or adequacy as a caption.

\end{document}